\documentclass[11pt]{article}
\usepackage{nlprs01}
\title{The Open Language Archives Community\\ and Asian Language Resources}
\author{
Steven Bird \\
\normalsize Linguistic Data Consortium \\
\normalsize University of Pennsylvania \\
\normalsize 3615 Market Street \\ 
\normalsize Philadelphia, PA 19104, USA \\ 
\normalsize {\sf sb@ldc.upenn.edu} \And
Gary Simons \\
\normalsize SIL International \\
\normalsize 7500 West Camp Wisdom Road \\
\normalsize Dallas, TX 75236, USA \\
\normalsize {\sf Gary\_Simons@sil.org} \And
Chu-Ren Huang \\
\normalsize Institute of Linguistics\\
\normalsize Academia Sinica\\
\normalsize 115 Nankang, Taipei, Taiwan\\
\normalsize {\sf churen@gate.sinica.edu.tw}
}
\usepackage{times,epsfig,boxedminipage,url,alltt}
\def\myurl#1{{[\small\url{#1}]}}
\def\smallurl#1{{\small\url{#1}}}
\def\scripturl#1{{\scriptsize\url{#1}}}

\def\elt#1{{\small\sf #1}}
\def\attr#1{{\small\sf #1}}
\def\code#1{{\small\sf #1}}

\def\OLAC{{\sc olac}}
\def\OAI{{\sc oai}}

\begin{document}
\maketitle

\begin{abstract}
The Open Language Archives Community (\OLAC) is a new
project to build a worldwide system of federated language
archives based on the Open Archives
Initiative and the Dublin Core Metadata Initiative.
This paper aims to disseminate the \OLAC\ vision to the language resources
community in Asia, and to show language technologists and linguists
how they can document
their tools and data in such a way that others can easily discover them.
We describe \OLAC\ and the \OLAC\ Metadata Set, then discuss two key
issues in the Asian context: language classification and multilingual
resource classification.
\end{abstract}

\section{Introduction}

Language technology and the linguistic sciences are
confronted with a vast array of \emph{language resources},
richly structured, large and diverse.
Multiple \emph{communities} depend on language resources, including
linguists, engineers, teachers and actual speakers.
Many individuals and institutions provide key pieces of the infrastructure,
including archivists, software developers, and publishers.
Today we have unprecedented opportunities to \emph{connect}
these communities to the language resources they need.

We can observe that the
individuals who use and create language resources
are looking for three things: data, tools, and advice.
By \emph{data} we mean any information that documents or describes a language,
such as a published monograph, a computer data file, or
even a shoebox full of hand-written index cards. The information could range
in content from unanalyzed sound recordings to fully transcribed and annotated
texts to a complete descriptive grammar. 
By \emph{tools} we mean computational resources that facilitate creating, viewing,
querying, or otherwise using language data. Tools include not just software
programs, but also the digital resources that the programs depend on, such as
fonts, stylesheets, and document type definitions.
By \emph{advice} we mean any information about
what data sources are reliable, what tools are appropriate in a given
situation, what practices to follow when creating new data, and so forth
(e.g. the Corpora List archives \myurl{http://www.hit.uib.no/corpora/}).
In the context of \OLAC, the term \emph{language resource} is broadly
construed to include all three of these: data, tools and advice.

\begin{figure}
\centerline{\includegraphics[width=\linewidth]{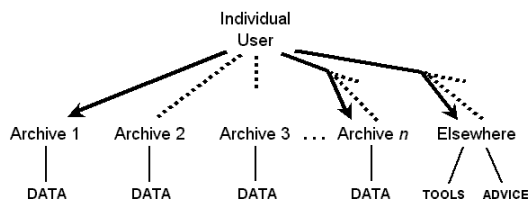}}
\caption{In reality the user can't always get there from here}
\label{fig:vision2}
\end{figure}
Unfortunately, today's user does not have ready access to the resources
that are needed. Figure~\ref{fig:vision2}
offers a diagrammatic view of this reality.
Some archives (e.g. Archive 1) do have a site on the internet which the user is
able to find, so the resources of that archive are accessible. Other archives
(e.g. Archive 2) are on the internet, so the user could access them in theory,
but the user has no idea they exist so they are not accessible in practice.
Still other archives (e.g. Archive 3) are not even on the internet. And there
are potentially hundreds of archives (e.g. Archive $n$) that the user
needs to know about. Tools and advice are out there as well, but are at many
different sites.

There are many other problems inherent
in the current situation. For instance, the user may not be able to find all
the existing data about a language of interest because different sites have
called it by different names (low \emph{recall}).
The user may be swamped with irrelevant resources because search terms
have important meanings in other domains (low \emph{precision}).
The user may not be able to use an accessible
data file for lack of being able to match it with the right tools. The user may
locate advice that seems relevant but have no basis for judging its merits.

As web-indexing technologies improve one might hope
that a general-purpose search engine should be sufficient to bridge the gap
between people and the resources they need.  However this is a vain hope.
First, many language resources, such as audio files
and software, are not text-based.  Second, many language names
have several variants, and these various strings regularly denote things
other than languages.  Third,
much of the material is not--and will never be--documented
in free prose on the web.
In place of traditional web-indexing, two new initiatives provide the
necessary infrastructure for language resource discovery.

The Dublin Core Metadata Initiative began in 1995 to develop
conventions for resource discovery on the web
\myurl{dublincore.org}.
The Dublin Core metadata elements represent a broad, interdisciplinary
consensus about
the core set of elements that are likely to be widely useful to support
resource discovery.  The Dublin Core consists of 15 metadata elements,
where each element is optional and repeatable: \elt{Title, Creator, Subject,
Description, Publisher, Contributor, Date, Type, Format, Identifier, Source,
Language, Relation, Coverage} and \elt{Rights}.
This set can be used to describe resources that
exist in digital or traditional formats.

The Open Archives Initiative (\OAI)
was launched in October 1999 to provide a common framework across
electronic preprint archives, and it has since been broadened
to include digital repositories of scholarly materials regardless
of their type.\footnote{\scripturl{www.openarchives.org}; \cite{LagozeVandeSompel01}}
To implement the \OAI\ infrastructure, an archive must comply
with two standards: the {\it \OAI\ Shared Metadata Set} (Dublin Core), which
facilitates interoperability across all repositories participating in the
\OAI, and the {\it \OAI\ Metadata Harvesting Protocol}, which allows
software services to query a repository using {\sc http} requests.

\begin{figure}
\centerline{\includegraphics[width=\linewidth]{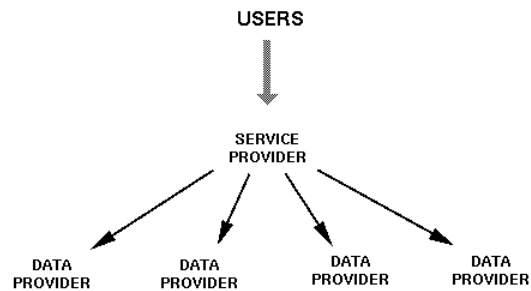}}
\caption{A Service Provider Accessing Multiple Data Providers}
\label{fig:white-paper2}
\end{figure}

\OAI\ archives are called ``data providers,'' and typically
have a submission procedure, a long-term storage system, and a
mechanism permitting users to obtain materials from the archive. An
\OAI\ ``service provider'' provides end-user services--such
as search functions over union catalogs--based on metadata harvested from
one or more data providers.  Figure~\ref{fig:white-paper2}
illustrates a
single service provider accessing three data providers
using the \OAI\ metadata harvesting protocol.
End-users only interact with service providers.

The \OAI\ infrastructure has the bottom-up, distributed character of the web,
while simultaneously having the efficient, structured
nature of a centralized database.  This combination is well-suited to
the language resource community, where the available data is growing
rapidly and where a large user-base is fairly consistent in how it describes
its resource needs.

\OAI\ data providers may support metadata standards in addition to the
Dublin Core.  Thus, a specialist community like the language resources
community can define a metadata format tailored to its domain.
Using the \OAI\ infrastructure, the community's
archives can be federated: a virtual meta-archive
collects all the information into a single place and end-users
can query multiple archives
simultaneously.  In the case of \OLAC, the Linguistic Data Consortium
has harvested the catalogs of ten participating archives and created a
search interface which permits queries over all 9,000+ records.
A single search typically returns records from multiple archives.
The prototype can be accessed via \myurl{www.language-archives.org}.

\section{A Core Metadata Set for Language Resources}
\label{sec:metadata}

The \OLAC\ Metadata Set extends the Dublin Core set only to
the minimum degree required to express those basic properties
of language resources which are useful as finding aids.
All fifteen Dublin Core elements are used in the \OLAC\ Metadata Set. In
order to suit the specific needs of the language resources community, the
elements have been ``qualified'' following principles articulated in
``Dublin Core Qualifiers'' \cite{DCQ00}, and explained below.
This section lists the \OLAC\ metadata elements and controlled vocabularies.
Full details are available from \myurl{www.language-archives.org}.

\subsection{Metadata attributes}

Three attributes -- \attr{refine}, \attr{code}, and \attr{lang} -- are used
throughout the metadata set to handle most qualifications to Dublin Core. Some
elements in the \OLAC\ Metadata Set use the \attr{refine} attribute to identify
element refinements. These qualifiers make the meaning of an element narrower
or more specific. A refined element shares the meaning of the unqualified
element, but with a more restricted scope \cite{DCQ00}.

Some elements in the \OLAC\ Metadata Set use the \attr{code} attribute to
hold metadata values that are taken from a specific encoding scheme. When an
element may take this attribute, the attribute value specifies a precise value
for the element taken from a controlled vocabulary or formal notation
(\S\ref{sec:cv}).
In such cases, the element content may also be used
to specify a free-form elaboration of the coded value.

Every element in the \OLAC\ Metadata Set may use the \attr{lang} attribute.
It specifies the language in which the text in the content of the element is
written. The value for the attribute comes from a controlled vocabulary
OLAC-Language.
By default, the \attr{lang} attribute has
the value ``en'', for English. Whenever the language of the element content is
other than English, the \attr{lang} attribute should be used to identify the
language. By using multiple instances of the metadata elements tagged for
different languages, data providers may offer their metadata records in
multiple languages.
Service providers may use this information in order to offer
multilingual views of the metadata.

\subsection{The elements of the OLAC Metadata Set}

In this section we present a synopsis of the elements of the \OLAC\ metadata
set.  Some elements are associated with a controlled vocabulary.  These are
parenthesized and discussed later.  Each element is optional and repeatable.

\begin{description}\setlength{\itemsep}{0pt}\setlength{\parskip}{0pt}

\item[\elt{Title}:]
A name given to the resource.

\item[\elt{Creator}:]
An entity primarily responsible for making the content of the resource
(OLAC-Role).

\item[\elt{Subject}:]
The topic of the content of the resource.

\item[\elt{Subject.language}:]
A language which the content of the resource describes or discusses
(OLAC-Language).

\item[\elt{Description}:]
An account of the content of the resource.

\item[\elt{Publisher}:]
An entity responsible for making the resource available.

\item[\elt{Contributor}:]
An entity responsible for making contributions to the content
of the resource (OLAC-Role).
      
\item[\elt{Date}:]
A date associated with an event in the life cycle of the resource.

\item[\elt{Type}:]
The nature or genre of the content of the resource (DC-Type).

\item[\elt{Type.linguistic}:]
The nature or genre of the content of the resource, from a linguistic
standpoint (OLAC-Linguistic-Type).

\item[\elt{Type.functionality}:]
The functionality of a software resource (OLAC-Functionality).

\item[\elt{Format}:]
The physical or digital manifestation of the resource
(OLAC-Format).

\item[\elt{Format.cpu}:]
The CPU required to use a software resource (OLAC-CPU).

\item[\elt{Format.encoding}:]
An encoded character set used by a digital resource (OLAC-Encoding).

\item[\elt{Format.markup}:]
The \OAI\ identifier for the definition of the markup format (OLAC-Markup).

\item[\elt{Format.os}:]
The operating system required to use a software resource (OLAC-OS).

\item[\elt{Format.sourcecode}:]
The programming language(s) of software distributed in source form
(OLAC-Sourcecode).

\item[\elt{Identifier}:]
An unambiguous reference to the resource within a given context
(e.g. URI, ISBN).

\item[\elt{Source}:]
A reference to a resource from which the present resource is derived.

\item[\elt{Language}:]
A language of the intellectual content of the resource (OLAC-Language).

\item[\elt{Relation}:]
A reference to a related resource.

\item[\elt{Coverage}:]
The extent or scope of the content of the resource
(e.g. spatial or temporal).

\item[\elt{Rights}:]
Information about rights held in and over the resource (OLAC-Rights).

\item[\elt{Rights.software}:]
Information about rights held in and over a software resource.
(OLAC-Software-Rights).

\end{description}

Observe that some elements, such as \elt{Format}, \elt{Format.encoding}
and \elt{Format.markup}
are applicable to software as well as to data.  Service providers can exploit
this feature to match data with appropriate software tools.

\subsection{The controlled vocabularies}
\label{sec:cv}

Controlled vocabularies are enumerations of legal values for the
\attr{code} and \attr{refine} attributes, and are currently
undergoing development.  In some cases, more than one value applies
and the corresponding element must be repeated, once for each
applicable value.  In other cases, no value is applicable and
the corresponding element is simply omitted.  In yet other cases, the
controlled vocabulary may fail to provide a suitable item, in which case
the most similar vocabulary item can be optionally specified,
and a prose comment included in the element content.

\begin{description}\setlength{\itemsep}{0pt}\setlength{\parskip}{0pt}
\item[OLAC-Language:]
A vocabulary for identifying the language(s) that the data is in, or that
a piece of linguistic description is about, or that a particular tool can
process.

\item[OLAC-Linguistic-Type:]
The primary linguistic descriptors for a language resource:
\code{transcription}, \code{annotation}, \code{description} and
\code{lexicon} (with subcodes for each type).

\item[OLAC-CPU:]
A vocabulary for identifying the CPU(s) for which the software is
available, in the case of binary distributions:
\code{x86}, \code{mips}, \code{alpha}, \code{ppc}, \code{sparc}, \code{680x0}.

\item[OLAC-Encoding:]
A vocabulary for identifying the character encoding used by a digital
resource, e.g. \code{iso-8859-1}, ...

\item[OLAC-Format:]
A vocabulary for identifying the manifestation of the resource.
The representation is inspired by MIME types, e.g. \code{text/sf} for
SIL standard format.  (\elt{Format.markup} is used to identify the particular
tagset.)  It may be necessary to add new types and subtypes to cover
non-digital holdings, such as manuscripts, microforms, and so forth
and we expect to be able to incorporate an existing vocabulary.

\item[OLAC-Functionality:]
A vocabulary for classifying the functionality of software,
again using the MIME style of representation, and using the
HLT Survey as a source of categories \cite{Cole97} as advocated
by the ACL/DFKI Natural Language Software Registry.  For example,
\code{written/OCR} would cover ``written language input, print or
handwriting optical character recognition.''

\item[OLAC-OS:]
A vocabulary for identifying the operating system(s) for which the software
is available:
\code{Unix}, \code{MacOS}, \code{OS2}, \code{MSDOS}, \code{MSWindows}.
Each of these has optional subtypes, e.g.
\code{Unix/Linux}, \code{MSWindows/winNT}.

\item[OLAC-Rights:]
A vocabulary for classifying the rights held over a resource, e.g.:
\code{open}, \code{restricted}, ...

\item[OLAC-Role:]
A vocabulary for identifying the role of a contributor or creator of the
resource, e.g.: \code{author}, \code{editor}, \code{translator},
\code{transcriber}, \code{sponsor}, ...

\item[OLAC-Software-Rights:]
A vocabulary for classifying the rights held over a resource, e.g.:
\code{open-source}, \code{royalty-free-library},
\code{royalty-free-binary}, \code{commercial}, ...

\item[OLAC-Sourcecode:]
A vocabulary for identifying the programming language(s) used by
software which is distributed in source form, e.g.:
\code{C++}, \code{Java}, \code{Python}, \code{Tcl}, \code{VB}, ...

\end{description}

\section{Issues for the Asian Language Resources Community}

Language identification is probably the most fundamental kind of
information that can be given to any language resource
\cite{Simons00}.  The most comprehensive knowledge base for language
identification is the Ethnologue \cite{Grimes00}, an online searchable
database which has been built up over fifty years.  The Ethnologue
database contains several types of information for each language: a
unique three-letter code, the country where this language
is spoken, alternative names, dialects, language classification,
comments, and references to the SIL bibliography.
This section discusses a variety of issues relating to
language identification and
to the specification of the languages covered by multilingual resources.
We believe these are two key issues for the Asian language resources
community.

\subsection{Issues with language identification}

In order to identify and catalog a language it is crucial to define
what counts as a language and to distinguish languages from dialects.
The editors of the Ethnologue have made thousands
of such decisions using advice from hundreds of experts around the world.
However, for many languages scholarship remains patchy
or else there is scholarly disagreement.
In such cases, the best that the Ethnologue can do is what it does
already--represent incomplete
knowledge and then produce periodic updates to reflect the results
of new research.

As \OLAC\ grows, Ethnologue codes will be deployed widely.  Each new
\OLAC-conformant archive will be faced with a range of issues in
seeking to associate language codes with language resources.
For concreteness, we have chosen for our examples the Formosan languages,
a group of Austronesian languages spoken in Taiwan.  We put ourselves
in the shoes of the field research group at Academia Sinica
(Elizabeth Zeitoun, personal communication)
and try to envisage the problems which they might
encounter in assigning Ethnologue codes to their language resources.

We see three broad categories of problem: over-splitting,
over-chunking and omission.  Over-splitting occurs when a
language variety is treated as a distinct language.
For example, Nataoran is given its own
language code (AIS) even though the scholars at
Academia Sinica consider it to be a dialect of Amis (ALV).
Over-chunking occurs when two distinct languages are treated as dialects
of a single language (there does not appear to be an example of this in
the Ethnologue's treatment of Formosan languages).
Omission occurs when a language is not listed. For example, two extinct
languages, Luilang and Quaquat, are not listed in the Ethnologue.
Another kind of omission problem occurs when the language is actually
listed, but the name by which the archivist knows it is not listed,
whether as a primary name or an alternate name. In such a case the
archivist cannot make the match to assign the proper code. For
instance, the language listed as Taroko (TRV) in the Ethnologue is known
as Seediq by the Academia Sinica; several of the alternate names listed
by the Ethnologue are similar, but none matches exactly.

Beyond these three problems with language identification,
a further type of problem concerns scholarly disagreement over
language family classification.
The Ethnologue follows the Oxford International
Encyclopedia of Linguistics \cite{Bright92} for most language families.
For the Austronesian languages, including the Formosan languages, the Ethnologue
follows the Comparative Austronesian Dictionary \cite{Tryon94}.
Additionally, some changes have been entered in the light
of more recent comparative studies.\footnote{
  More information is available online at
  \scripturl{http://www.ethnologue.com/ethno_docs/introduction.asp}.
  The Formosan language family can be viewed at
  \scripturl{http://www.ethnologue.com/show_family.asp?subid=982}.}
Academia Sinica has developed its own language family classification
scheme for Formosan languages, and this differs from the Ethnologue/Tryon
scheme.  Additionally, languages typically have many variant names, and
scholars may disagree on the choice of a canonical
name for the language.  For example,
the Academia Sinica scholars believe Taroko to be a variant of Seediq, while
the Ethnologue/Tryon would presumably consider Seediq to be a variant of Taroko.

The consequences of these problems for classification and
retrieval are obvious.  In the case of over-splitting, as with
AIS and ALV mentioned above, someone
searching for Amis resources will need to know to search
over both codes.  An archivist
cataloging a resource which is ambiguous with respect to the AIS/ALV
distinction (perhaps because it was created by someone who
did not believe in the distinction) may need to assign both codes.
In the case of over-chunking, an archivist cannot specify
the individual language but must use a code which designates two or more
languages.  Someone searching for resources in one of those
languages will experience
lower precision.  In the case of omission, no language code can be assigned,
and classification and search must fall back to using conventional string
representations for language names (with the attendant precision and
recall problems).  In the case of differing language family classifications, the
precision and recall of searches on language family names are reduced.

All of these problems can be addressed through existing Ethnologue
mechanisms.\footnote{See the four questionnaires at
\scripturl{http://www.ethnologue.com/ethno_docs/questionnaires.asp}.}
However, \OLAC\ metadata and service providers could offer
complementary remedies.

{\bf Controlling element content.}
The \elt{Language} and \elt{Subject.language} elements permit the language
code to be specified in the \attr{code} attribute, while the element content
is unrestricted.  A community of Formosan scholars could develop a controlled
vocabulary for identifying speech varieties down to any level of detail they
liked, and then use those terms as the content of the \elt{Language} or
\elt{Subject.language} element.  For example, the following are five
varieties of the Bunun language:

{\scriptsize
\begin{verbatim}
<language code="x-sil-BNN">Northern/Takituduh</>
<language code="x-sil-BNN">Northern/Takibakha</>
<language code="x-sil-BNN">Central/Takbanuaz</>
<language code="x-sil-BNN">Central/Takivatan</>
<language code="x-sil-BNN">Southern/Isbukun</>
\end{verbatim}
}

If no Ethnologue code corresponded to the group of languages
in question, as in the Amis/Nataoran case, the code attribute
could be omitted (though this would
prevent recall on the Ethnologue code).  This general approach
could be formalized by permitting subcommunities to register an
encoding scheme as a controlled vocabulary with a unique
name.  That name would be specified as the value of a new
\attr{scheme} attribute,
and the element content would be constrained to be an item from
the corresponding vocabulary.  These approaches would
address the problems of over-chunking and omission.

{\bf Registering language groups with an OLAC registration service.}
While the classification of a language is sometimes treated as metadata
for resources in that language, we believe that a more appropriate location
for this type of finding aid is in \OLAC\ service providers.
\OLAC\ could maintain a language classification server which would
house a comprehensive list of language family names
and their extensional definitions (i.e. sets of Ethnologue codes).  The
server would permit users to define their own language group names or their
own versions of existing group names.  For instance,
Academia Sinica could register a language group name
\code{AS:Amis} with the extension \{\code{ALV}, \code{AIS}\}.
Searching on their notion of ``Amis'' would return resources classified
under both codes.
Entire classification schemes with complex hierarchies could be
represented in this fashion.
\OLAC\ service providers could index their harvested metadata using these names,
allowing any user to perform searches using any classification
scheme.  Over time, the more respected and popular classifications
could be identified and accorded due prominence.  This mechanism
would address the problems of over-splitting and differing
classification.

\subsection{Issues with multilingual resources}

For many language resources it is necessary to identify more than one language.
In some cases, such as MT systems and bilingual lexicons, there is an
added complication, namely \emph{directionality}.  Someone looking for such
a resource will usually want to specify source and/or target languages.
For example, in searching for a Korean-to-English resource, the user would not usually
be interested in discovering English-to-Korean resources.
Thus, there is an \emph{a priori} need for resources to be described using
\emph{ordered pairs} of languages.

\OLAC\ provides some mechanisms which can be applied to these cases.
First, \OLAC\ metadata elements are repeatable, e.g.~a
system which can process multiple languages will be described using multiple
\elt{Subject.language} elements, one per language.
Second, \OLAC\ incorporates
directionality in its distinction between \elt{Language} and \elt{Subject.language}.
Are these simple mechanisms adequate for the convenient and accurate
description and discovery of multilingual resources?
In this section we enumerate the main
types of multilingual resource and show how \OLAC\ metadata can be used to classify
them.  We report some problems and discuss possible solutions.

\subsubsection*{Machine Translation Systems}

The simplest type of MT system is a unidirectional
system which translates language $S$ to language $T$.  Here, the intended audience
of such a system is assumed to be the
speakers of language $T$ who use the system to
access documents in another language $S$.
The \OLAC\ solution would be to designate $S$ as the \elt{Subject.language}
and $T$ as the \elt{Language}.

Note that ``audience'' is slightly problematic.
Such an MT system may be intended for an audience of $S$ speakers who
wish to translate their documents into language $T$.
The problem here is not with directionality
but with the notion of ``audience'' in the \OLAC\ definition of \elt{Language}.
The definition could be adjusted to remove this problem.

Next in order of complexity is the bidirectional case, where a system translates
in both directions between languages $X$ and $Y$.  Extending the previous
solution, we would
designate both $X$ and $Y$ as \elt{Language} and \elt{Subject.language}.  Ideally,
we would use order or structure to group the languages appropriately:

\pagebreak
\begin{alltt}\small
<pair><Subject.language code={\it X}/>
      <Language code={\it Y}/></pair>
<pair><Subject.language code={\it Y}/>
      <Language code={\it X}/></pair>
\end{alltt}

However, \OLAC\ metadata is flat and unordered.  The only available
options are permutations of the following, in which we can make no
contrastive use of order.

\begin{alltt}\small
<Language code={\it X}/>
<Language code={\it Y}/>
<Subject.language code={\it X}/>
<Subject.language code={\it Y}/>
\end{alltt}

Although this loses information, we do not
believe it presents a problem for typical kinds of retrieval.
Queries for an MT system (i) from $X$; (ii) from $Y$;
(iii) to $X$; (iv) to $Y$; (v) from $X$ to $Y$; or (vi) from $Y$ to $X$,
will discover the system described above.

Next are MT systems which translate from one language into many, or
from many languages into one (star configurations).
Here the obvious approach is adequate:

\begin{alltt}\small
{\it One-to-many:}
<Subject.language code={\it S}/>
<Language code={\it T1}/>
<Language code={\it T2}/>
<Language code={\it T3}/>

{\it Many-to-one:}
<Subject.language code={\it S1}/>
<Subject.language code={\it S2}/>
<Subject.language code={\it S3}/>
<Language code={\it T}/>
\end{alltt}

Finally, there are MT systems which translate \emph{from} and \emph{to}
all languages in a set of $n$ languages.
Here again the obvious approach is adequate, and is clearly
superior to a solution where all \mbox{$n(n-1)$} ordered pairs are enumerated.

\begin{alltt}\small
<Subject.language code={\it X}/>
<Subject.language code={\it Y}/>
<Subject.language code={\it Z}/>
<Language code={\it X}/>
<Language code={\it Y}/>
<Language code={\it Z}/>
\end{alltt}

\subsubsection*{Multilingual Lexicons}

A unidirectional bilingual lexicon
with lemmas in language $S$ and definitions in language $T$ is
described like a unidirectional MT system.  The \OLAC\ solution is to
designate $S$ as the \elt{Subject.language} and $T$ as the
\elt{Language}.
A fully bidirectional lexicon intended for use by speakers of either
language would be described in the same fashion
as a bidirectional MT system:

\begin{alltt}\small
<Language code={\it X}/>
<Language code={\it Y}/>
<Subject.language code={\it X}/>
<Subject.language code={\it Y}/>
\end{alltt}

As before, this solution has lost some structure, but retrieval behavior
is correct nonetheless.  Of course, this metadata could equally be the
collection-level metadata for a set of two monolingual dictionaries,
one in language $X$ and one in language $Y$.  However,
the \elt{Type.linguistic} element would distinguish these cases by
having different vocabulary items for monolingual and bilingual lexicons.

Multilingual lexicons may also exhibit star configurations:
one-to-many (lexicons with definitions in multiple languages); or
many-to-one (comparative wordlists).  Here the treatment is analogous
to the corresponding MT systems discussed above.

Finally, multilingual lexicons may map between all pairs of a set of languages,
as in the case of some termbanks.  In this case, all the languages are designated
both using \elt{Language} and \elt{Subject.language}.

\subsubsection*{Text Collections}

The most simple case of a text collection is a set of texts in a single
language.  Usually, the language in these texts would be
described using the \elt{Language} element.  However, in the situation
where the text collection is intended to document a language, then it
is simultaneously \emph{in} and \emph{about} that language.
Accordingly, it would be tempting to describe this with both the \elt{Language} and
\elt{Subject.language} elements.  Rather than let the decision about metadata
depend on the intent of the creator of the resource (which may not be known),
or on the typical usage of the resource (which may change through time),
we think it would be simplest to describe such situations using the
\elt{Language} element only.  This approach generalizes to the case of a
text collection spanning multiple languages (e.g. where each text comes from
one of the Formosan languages).  Here, collection-level
metadata would provide a \elt{Language} element for each language
represented in the collection.

In the case of bilingual aligned texts (bitexts) there is normally a
directionality, since one of the texts is primary and the other is a
translation.  Here, we specify the primary language using
\elt{Subject.language}, and the translation using \elt{Language}.

A complication arises when the texts or bitexts employ analytical
notations.  For a text in language $X$ transcribed in a notation
(such as the International Phonetic Alphabet) which is
inaccessible to speakers of $X$ (assuming any exist), it would
make sense to use \elt{Subject.language} instead of \elt{Language}.
The situation becomes more vexing for texts with embedded annotations
(such as the notation of Conversation Analysis), where non-specialists
could discard the annotations to get a conventional text.  Here, the
use of both \elt{Subject.language} and \elt{Language} seems to be
indicated.

\subsubsection*{Descriptions}

The final category we consider is linguistic descriptions, such
as field notes and grammars.  In the usual case, the language
described is specified with \elt{Subject.language} while the language
of the commentary is specified with \elt{Language}.  One interesting
case is where a third language is used for elicitation.  For example,
a sentence from Amis may have been elicited using a sentence
from Chinese, and both sentences may have been entered in the field notes.
Next, commentary in the language of the linguist,
such as English, may have been added.
In this case, we would say that the field notes include bitext,
and the languages of the bitext would be described in the usual way.
The audience language of the field notes would also be specified.
Using \OLAC's flat metadata, we would specify the languages as follows:

\begin{alltt}\small
<Subject.language code="x-sil-AIS"/>
<Language code="x-sil-CFR"/>
<Language code="x-sil-ENG"/>
\end{alltt}

We believe this is perfectly adequate for the majority of retrieval purposes.
If it were necessary to represent the structure more accurately,
\emph{two} \OLAC\ records could be associated with the same resource,
one describing the field notes as a whole (with the above language
designations), and one describing the bitext content (with just the
AIS and CFR designations).  The different linguistic types would be
expressed using the \elt{Type.linguistic} element, and the two records
could refer to each other using the \elt{Relation} element:

\begin{alltt}\small
<Relation refine="isPartOf">{\it id1}</Relation>
<Relation refine="hasPart">{\it id2}</Relation>
\end{alltt}

\section{Conclusions}

This paper has presented the leading ideas of the
Open Language Archives Community (\OLAC), along with
its metadata set and controlled vocabularies.  Language
resources exhibit great diversity, and include all types
of data, tools and advice.  As collections
of language resources proliferate, \OLAC\ will make it
maximally easy for members of the language resources community to
discover each other's resources.  Another dimension
of \OLAC, not discussed here, will permit community-agreed best
practices to be identified, greatly facilitating
resource re-use.

Members of the Asian language resources
community are encouraged to join \OLAC\ and contribute to
the development of \OLAC\ metadata, vocabularies, and archives.

\section*{Acknowledgments}

\small

The work reported here is supported by the National Science
Foundation under grants:
9910603 \emph{International Standards in Language Engineering},
9978056 \emph{TalkBank},
9983258 \emph{Linguistic Exploration}, and
by a Taiwanese National Digital Archives Pilot project
\emph{Chinese and Austronesian Corpora}.  The LDC prototype
was developed by \'Eva B\'anik.

\raggedright\small
\bibliographystyle{acl}

\def\vs{\vspace*{-1ex}}

\end{document}